\pdfoutput=1

\documentclass[a4paper,fleqn]{cas-dc}

\usepackage[numbers,authoryear]{natbib}

\usepackage{setspace}
\usepackage[ruled,lined,boxed,commentsnumbered]{algorithm2e} 
\usepackage{multirow}
\usepackage{booktabs}
\usepackage{bm}
\usepackage{newfloat}
\usepackage{listings}
\usepackage{makecell}
\usepackage{hyperref}
\usepackage{comment}
\usepackage{amsmath}
\usepackage{mathtools}

\usepackage{algorithmic}
\usepackage{caption}
\def\tsc#1{\csdef{#1}{\textsc{\lowercase{#1}}\xspace}}
\tsc{WGM}
\tsc{QE}
\tsc{EP}
\tsc{PMS}
\tsc{BEC}
\tsc{DE}

\newcommand{\y}{\mathbf{y}}

\newcommand{\YYG}[1]{{\color{black} #1}}
\newcommand{\GRT}[1]{{\color{black} #1}}

\def\ie{\emph{i.e.}}
\def\eg{\emph{e.g.}}
\def\wrt{\emph{w.r.t.~}}

\def\ours{DecomCAM}

\def\paperTitle{DecomCAM: Advancing Beyond Saliency Maps through Decomposition and Integration}

\UseRawInputEncoding

\begin{document}
\let\WriteBookmarks\relax
\def\floatpagepagefraction{1}
\def\textpagefraction{.001}

\shorttitle{\paperTitle}

\shortauthors{Yuguang Yang et~al.}

\title [mode = title]{\paperTitle}                      
\tnotemark[1]


\author[1]{Yuguang Yang}
[style=chinese, orcid=0000-0002-6394-4645]
\fnmark[1]

\author[1]{Runtang Guo}[style=chinese]
\fnmark[1]

\author[1]{Sheng Wu}[style=chinese]
\author[1]{Yimi Wang}[style=chinese]

\author [5]{Linlin Yang}[style=chinese]

\author [4]{Bo Fan}[style=chinese]  \cormark[1]
\ead{oceanpearl@126.com}

\author [4]{Jilong Zhong}[style=chinese]


\author [1,2]{Juan Zhang}[style=chinese]

\author [1,2]{Baochang Zhang}[style=chinese]


\affiliation[1]{organization={Beihang University},
    city={Beijing},
    country={China}
    }

\affiliation[2]{
organization={Zhongguancun Laboratory},
city={Beijing},
country={China}
}

\affiliation[4]{organization={Academy of Military Science Defense Innovation Institute},
    city={Beijing},
    country={China}
    }

\affiliation[5]{organization={Communication University of China},
    addressline={State Key Laboratory of Media Convergence and Communication}, 
    city={Beijing},
    country={China}
    }

\fntext[fn1]{These authors have contributed equally to this work and share first authorship.}
\cortext[cor1]{Corresponding authors.}

\begin{abstract}
Interpreting complex deep networks, notably pre-trained vision-language models (VLMs), is a formidable challenge.
Current Class Activation Map (CAM) methods highlight regions revealing the model's decision-making basis but lack clear saliency maps and detailed interpretability.
To bridge this gap, we propose \ours, a novel decomposition-and-integration method that distills shared patterns from channel activation maps. Utilizing singular value decomposition, \ours~decomposes class-discriminative activation maps into orthogonal sub-saliency maps (OSSMs), which are then integrated together based on their contribution to the target concept. 
Extensive experiments on six benchmarks reveal that \ours~not only excels in locating accuracy but also achieves an optimizing balance between interpretability and computational efficiency. Further analysis unveils that OSSMs correlate with discernible object components, facilitating a granular understanding of the model's reasoning. This positions \ours~as a potential tool for fine-grained interpretation of advanced deep learning models. The code is avaible at \url{https://github.com/CapricornGuang/DecomCAM}.
\end{abstract}


\maketitle

\section{Introduction}

Deep convolutional neural networks (CNNs) have achieved remarkable success across a variety of visual tasks, including object detection \citep{bib3}, saliency detection~\citep{fang2022densely}, and image retrieval~\citep{bib4}. Recently, the advent of pre-trained vision-language models (VLMs) \eg, CLIP~\citep{radford2021learning}, have signified a leap forward, learning generalized visual representations that grasp a broad spectrum of visual concepts that can be aligned with textual descriptions. This capability enables VLMs to make novel few-shot or zero-shot decisions in an open-world setting \citep{guo2023mvp, yang2023self, zhang2022tip}. However, the very breadth of these concepts, while beneficial for generalization, also raises concerns about interpretability. What concepts contribute to the final prediction? Can we produce accurate visual explanations for these concepts? These questions serve as motivation for us to develop post-hoc methods to interpret.

\begin{figure}[t]
\centering
\includegraphics[width=\linewidth]{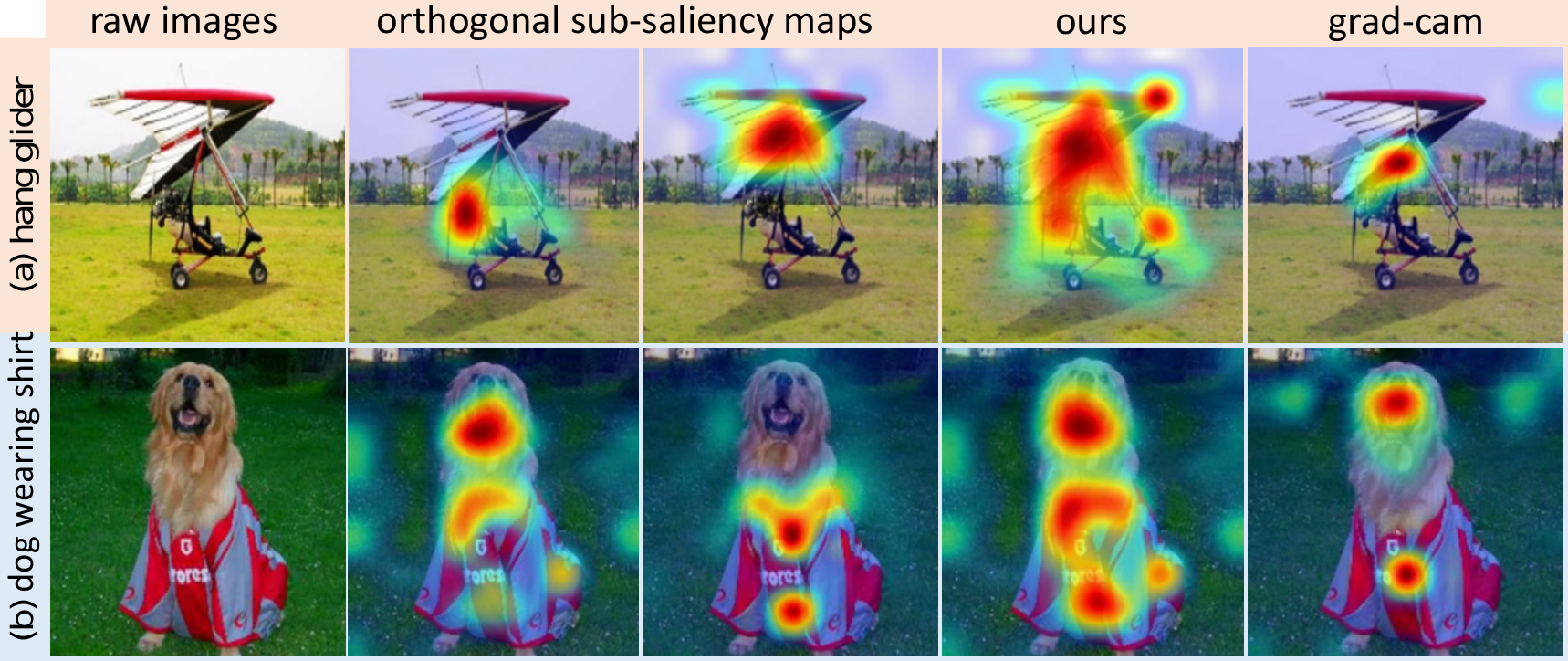}
\caption{Illustration of our DecomCAM method in action. We present the orthogonal sub-saliency maps produced by \ours, which isolate distinct attributes within the images. This capability enables our method to handle complex concepts, \eg~``dog wearing a shirt'', resulting in a comprehensive and clear saliency map that captures detailed aspects of the concept to be interpreted.}
\label{fig:teaser}
\end{figure}

Class Activation Map techniques (CAMs) provide an effective pathway by illuminating the salient regions within an input image that contribute most strongly to the model's final prediction~\citep{selvaraju2017grad,bib24,bib35,bib36,bib43,bib29,bib30,bib37}.
As shown in the last column of Fig.~\ref{fig:teaser}, the highlighted regions indicate the model's decision basis when predicting the target concept.
These maps acts as an unsupervised object detector \citep{bib24, wagner2019interpretable, Bi-CAM}, locating the regions within the input image that contribute most significantly to the final concept. \YYG{Specifically, CAM \citep{bib22} aggregates the feature maps of the last convolutional layer using a weighted average to generate a semantic-aware saliency map. These weights are obtained from a global average pooling layer \citep{lin2013network}, which effectively captures the importance of each feature map about the target concept. Building upon this foundation, several variants have emerged to enhance the flexibility and accuracy of saliency maps. These models adapt CAM to interpret any layer of the deep model without a retraining process using the back-propagation gradients, \eg~GradCAM, GradCAM++, \textit{etc.} \citep{selvaraju2017grad,bib24, bib26}, or final confidence values obtained from forward-propagation, \eg~ScoreCAM, EigenCAM, \textit{etc.} \citep{bib29,bib30, bib37, bib43, muhammad2020eigen}, to assess the weight of each activation map.
 Channel-wise linear aggregation, as employed in CAM and its variants, is pivotal for highlighting the most influential feature combinations relevant to the target concept. Consequently, the quality of saliency maps hinges on both the fidelity of activation maps and the chosen aggregation method.
}

\begin{figure*}[tp]
\centering
\vspace{-\abovecaptionskip} 
\includegraphics[width=\textwidth]{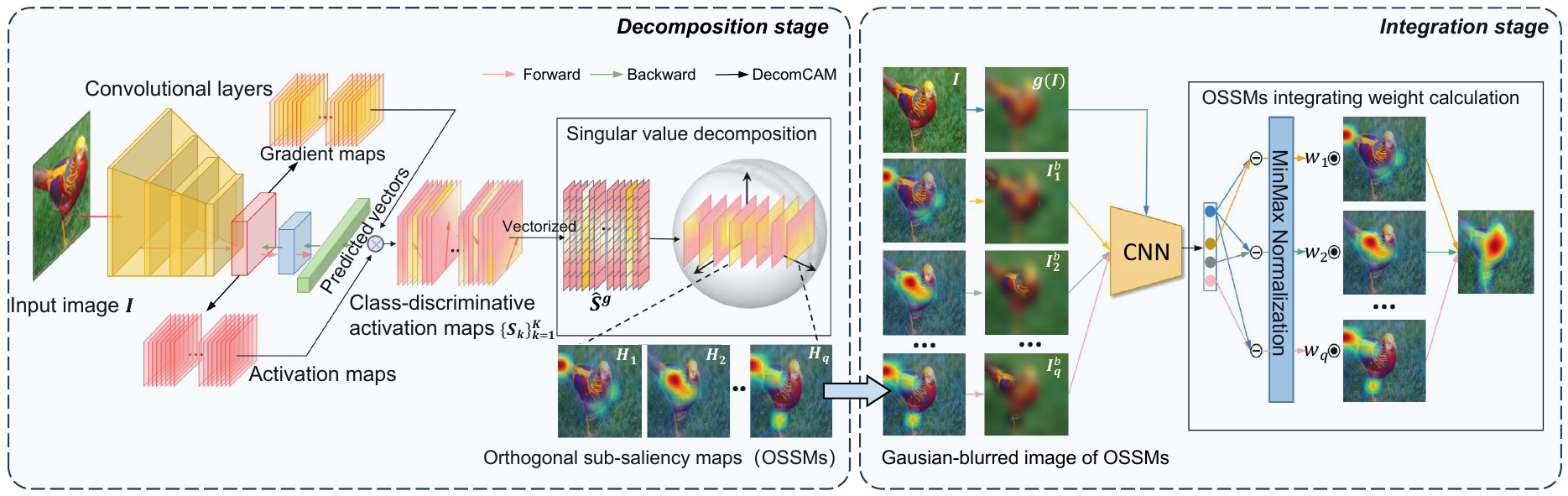}
\caption{ \ours~overview.  The proposed method operates through a streamlined two-stage process: a) In the decomposition stage, \ours~utilizes the gradient maps derived from predicted vectors and activation maps to create the class-discriminative activation maps, $\hat{S}^{g}$. Applying singular value decomposition to $\hat{S}^{g}$, we select the top-$Q$ Orthogonal Sub-Saliency Maps (OSSMs) that represent the top-$Q$ common activation patterns across different channels. These OSSMs indicate detailed visual features related to the target concept. b) In the integration stage, a Gaussian-blurred version of the input image and OSSMs are processed through the CNN to assess their respective impacts on the model's confidence. The difference in confidence levels informs the aggregation weights of OSSMs, which are then used to integrate OSSMs together to generate a clear and comprehensive saliency map.  
}
\vspace{-\abovecaptionskip} 
\label{fig:overview}
\end{figure*}

However, with the emerging number of channels, which often scale to thousands in recent VLM models, two significant concerns raise: 
1) Noise issue. 
The advancement of VLMs has witnessed the expanding parameter and data scale of deep learning models, resulting in activation maps containing a richer set of visual concepts~\citep{shao2023textual, barraco2022unreasonable, wang2023efficient}. \YYG{Although such richness improves model performance, it presents challenges for reliable visual interpretation using saliency maps, as noise may be hidden within the activation maps and must be carefully considered in model interpretations. For instance, when interpreting fine-grained targets such as the eye of a cat, other channels may contain information about the cat's body, ears, or tail, adding to the complexity of interpretation. Moreover, gradient-based methods like GradCAM \citep{selvaraju2017grad} and GradCAM++ \citep{bib24} may exacerbate the noise issue in saliency maps due to problems such as gradient saturation or vanishing~\citep{bib36, bib43}.}

2) Significant factors entanglement issue. 
Existing methods~\citep{bib24, bib26,bib29, bib14, bib30} typically employ a weighted aggregation operation on the activation map to produce the final saliency map. \YYG{However, this direct aggregation approach lacks a statistical analysis, which may lead to the entanglement of significant factors learned by the model. For example, when identifying a bird, important features like the head, wings, and beak may each play a distinct role. However, during aggregation, these factors may be mixed together. Consequently, it provides limited insights into the model's decision-making process and fails to capture the interactions among different channels. Although methods like EigenCAM \citep{cheng2023deeply}, EigenGradCAM \citep{jacobgilpytorchcam} attempt to decompose the activation map to extract these significant factors, they fail to achieve a reliable target-specific interpretation.}

To address these two issues, this paper proposes a two-stage interpretability method, Decomposition-and-Integration Class Activation Map (\ours). \YYG{\ours~can analyze the correlation between each channel and extract dominant principal components using Singular Value Decomposition (SVD). Compared with the aggregation scheme in existing CAM techniques, SVD can effectively decompose the intrinsic elements from activation maps while reduce the noise effect.} The conceptual overview of this approach is depicted in Fig.~\ref{fig:overview}. 
Specifically, the decomposition phase begins by constructing class-discriminative activation maps through the multiplication of the activation map and the gradient maps derived from the target concept. 
But given the potential channel redundancy issue, we additionally introduce a gradient-guided channel selection process to pick out the top-$P$ significant channels. \YYG{Specifically, we compute the pixel-wise sum of each channel's gradient map and select the top $P$ channels with the highest sums, thereby filtering out less important channels and reducing noises.} Following this selection, SVD is applied to the refined class-discriminative activation maps, breaking them down into distinct orthogonal sub-saliency maps (OSSMs). These decomposed components allow for a clearer attribution of model decisions to specific visual features.
By selecting the OSSMs corresponding to the top-${Q}$ singular values, we can preserve the majority of the variance information from the original activation maps and reduce redundancy. These ${Q}$ OSSMs capture top-${Q}$ common activation patterns across different channels. 
\YYG{Additionally, due to the tendency of noise-prone channels to contribute less variance information, they occupy smaller singular value eigenvectors. Consequently, the OSSMs naturally aggregated by the SVD method not only represent visual concepts but also contribute to smoothing the saliency map, thereby reducing noise.}
As shown in the ``orthogonal sub-saliency maps'' column of Fig.~\ref{fig:teaser}, we see that each OSSM demonstrate a clear heatmap that closely related to the visual concepts. Specially, for ``dog wearing shirt'', the proposed OSSMs highlight the regions of ``dog'' and ``shirt''. 
In the integration stage, we fuse these OSSMs based on their contribution to the final prediction. This further balances the weights of OSSMs and achieves better visualization. \YYG{To avoid the false confidence issue of generating final saliency maps by a direct summation of OSSMs ~\citep{bib29}, we propose to leverage the impact of OSSM's forward passing score on the target concept as their weights. Specifically, we begin by applying Gaussian blur to the original image to create a reference image. Then, we enhance the regions highlighted by OSSMs in the reference image. We determine the weights by comparing the final scores obtained from forward inference using these two images. The integration stage of Fig.~\ref{fig:overview} clearly demonstrate this process. It's worth noting that while similar operations exist in ScoreCAM, gScoreCAM \cite{chen2022gscorecam, bib30}, our approach represents the first attempt to utilize this method to recombine decomposed sub-saliency maps.
} Fig.~\ref{fig:teaser} (b) ``ours'' column shows our final saliency maps almost highlight all the visual components that aligned with human decision making.


To evaluate \ours, we conduct extensive experiments covering three critical dimensions of interpretability: locating interpretation, class-discriminative causal interpretation, and attribute analysis. These experiments are carried out on six benchmarks for the CLIP model. Our results demonstrate \ours's superiority in locating and interpreting ability compared to the prior art. Our analysis reveals \ours's well balance between interpretability and computational efficiency. Furthermore, we conduct the attribute analysis experiment to show that the OSSMs display a strong correlation with the interpretable visual concepts, thus paving an avenue for fine-grained interpretation for deep models. In summary, the contributions of our method are as follows:
\begin{itemize}
\item We propose \ours, a novel decomposition-and-integration approach for interpretability, which simultaneously provides interpretable sub-saliency maps and an overall saliency map in response to the target concept.
\item In the decomposition phase, we extract orthogonal sub-saliency maps (OSSMs) from class activation maps via singular value decomposition, leading to more discernible interpretative elements. In the integration phase, we further calculate the aggregation weights of these OSSMs with their contribution to the target concept for a comprehensive interpretation.

\item  We present \ours's locating ability using CLIP on multiple datasets, including ImageNetV2, MSCOCO 2017, PartImageNet, and PASCAL VOC 2012, along with its causal interpretability assessed on the specially constructed PS-ImageNet. Our experimental results illustrate that DecomCAM can efficiently and precisely localize or interpret both objects and objects' parts, surpassing the prior art by a large margin. 

\item We further conduct attribute analysis of the yielded OSSMs using CLIP on PASCAL Part datasets. The results demonstrate that our OSSMs are strongly aligned with visual concepts. This confirms that our OSSMs contribute to a more detailed understanding of the model's decision-making process, highlighting the potential for fine-grained interpretability.

\end{itemize}

\section{Related Work}
\subsection{Interpretation with CAM}
Interpretation methods target understanding and explaining the predictions made by deep models and can be classified into occlusion-based~\citep{petsiuk2021black, petsiuk2018rise}, perturbation-based~\citep{bib18, bib20, bib21, duan2023bandit} and CAM-based~\citep{bib22,selvaraju2017grad,bib30,bib37}. CAM-based methods perform a linear combination of activation maps and class-discriminative channel weights to generate saliency maps. We focus on CAM-based methods because they are based on the localization property of convolution, and easy to interpret spatial information. 

Existing CAM-based works~\citep{bib22,selvaraju2017grad,bib30,bib37} focus on the estimation of class-discriminative channel weights. The original CAM method employs the weights of the global average layer to derive the class-discriminative weights of the final classification layer~\citep{bib22}. However, it requires replacing the final layer with a global average layer and re-training the model. Instead, Grad-CAM and its follow-up~\citep{selvaraju2017grad,bib24,bib35} avoid re-training by utilizing gradients to obtain class-discriminative weights for each channel.
However, the gradient produced by backpropagation suffers from gradient saturation~\citep{bib36} and or gradient vanishing~\citep{bib43}, leading to unfaithful spatial explanations. In this case, gradient-free CAM methods like~\citep{bib36,bib43,bib29,bib30,bib37} determine the importance of feature maps \wrt the target via removing the information related to feature maps and observing the impact on the network's output. 
\YYG{
However, these works focus on the channel-wise spatial activations and their class-discriminative channel weights, without considering the connection across channels. Instead, DecomCAM incorporates the concept of ``decomposition''  for CAM, which decomposes and then integrates across channels for both activations and weights. This will help remove the noise of the final saliency map and discover common activation patterns for the target.
}

\subsection{Interpretation with Decomposition}

Decomposition is a fundamental tool for analyzing data and features. It can reveal common patterns~\citep{collins2018deep,muhammad2020eigen}, simplify complex data into interpretable components~\citep{cheng2023deeply,zhou2018interpretable,praggastis2022svd}, and reduce noise~\citep{muhammad2020eigen}.

To discover the common patterns, \cite{muhammad2020eigen} and \cite{collins2018deep} adopt SVD for different tasks. Specifically, EigenCAM \citep{muhammad2020eigen} computes and visualizes the principle components of features from the convolutional layers, while~\cite{collins2018deep} discover semantically matching regions across a set of images by decomposing their activations of a pre-trained neural network. Similar to \citep{muhammad2020eigen}, \cite{praggastis2022svd} also propose to decompose features from the convolutional layers. \YYG{DecomCAM takes a further step by directly applying the results of decomposition for model interpretation. By identifying OSSMs, DecomCAM provides a more direct and interpretable representation of the model's decision-making process, enabling a deeper understanding of the underlying semantics.}

Other works~\citep{cheng2023deeply,zhou2018interpretable} typically use decomposition to get semantic components. 
\cite{cheng2023deeply} focuses on bridging the semantic gap between higher and lower network layers by decomposing key features to trace supporting evidence through layer perturbation. \cite{zhou2018interpretable} employs decomposition in conjunction with least-squares optimization to align activation maps with semantically interpretable components derived from an extensive concept corpus. 
 \YYG{However, these approaches differ from DecomCAM in their utilization of decomposed components for model interpretation. While they focus on obtaining semantic components, they do not directly apply the results of decomposition for interpreting model decisions. Additionally, they lack an in-depth analysis of the semantic representation power of the decomposed components, a feature that is central to DecomCAM methodology.}

\section{Methodology}

~\label{sec:methodology}
The aim of DecomCAM is to achieve more interpretable components and better final visualization based on class discriminative activation so that we can reveal the common patterns of discriminative image regions. In this section, we first present GradCAM~\citep{selvaraju2017grad} as our preliminary in Sec.~\ref{subsec:gradcam}. We then introduce the decomposition 
and integration phases of our DecomCAM in Sec.~\ref{subsec:decomcam}. The illustration of decomposition and integration can be found in Fig.~\ref{fig:overview} (a) and~(b), respectively. Last, we present in Sec.~\ref{subsec:app} how DecomCAM can be applied for the VLMs-based method, CLIP~\citep{radford2021learning}, to show the potential to discover meaningful visual concepts.

\subsection{GradCAM}~\label{subsec:gradcam}
Given an input image $\bm{I} \in \mathbb{R}^{3 \times H \times W}$ and a deep convolutional network ${f}:\bm{I} \mapsto \y$, where $\y \in \mathbb{R}^{C}$ denotes the classifier's prediction for C categories. 
For GradCAM, $y^c$ represents the score before the softmax for the c-th class, and $\bm{A}_k \in \mathbb{R}^{M \times N}$ denotes the activation map of the k-th channel of a convolutional layer, typically referring to the last convolutional layer. Consequently, the gradient of the score $y^c$ with respect to $\bm{A}_k$ can be expressed as $\frac{\partial y^c}{\partial \bm{A}_k}$. The importance weights of neurons $\alpha^c_k$  are calculated by performing global average pooling on the gradients of the corresponding activation maps across the width and height dimensions:

\begin{equation}
\label{eq:gradcam_channel}
\begin{aligned}
\bm{S}_k =  (\frac{1}{{Z}} \sum_{i=1}^{M} \sum_{j=1}^{N} (\frac{\partial y^c}{\partial \bm{A}_k})_{ij} )\odot \bm{A}_k,
\end{aligned}
\end{equation}
where $\frac{\partial y^c}{\partial \bm{A}_k}$ denotes the gradient map of each channel, and $\sum_{i=1}^{M} \sum_{j=1}^{N} (\frac{\partial y^c}{\partial \bm{A}_k})_{ij} $ estimates the importance value of each channel to the final prediction $y^c$ calculated with back-propagation, and $Z$ is the normalized coefficient.
With the weighted combination of forward activation maps, and follow by a ReLU to obtaion, the saliency map of GradCAM $L_\text{GradCAM}$ is performed as:
\begin{equation}
\label{eq:gradcam_fusion}
\begin{aligned}
L_\text{GradCAM} =  \text{ReLU}(\sum_{k=1}^K \bm{S}_k).
\end{aligned}
\end{equation}

This results in a coarse heatmap of the same size as the convolutional feature maps, and can be upsampled to the input image resolution using bilinear interpolation. The weights $\alpha^c_k$ is used to capture the 'importance' of feature map k for a target class c. The ReLU applys to the linear combination of maps highlights the features that have positive contribution in increasing $y^c$, whereas the negtive pixels have proned to have adverse influnence in localization.

\subsection{DecomCAM}~\label{subsec:decomcam}

\noindent \textbf{{Decomposition.}} Based on Eq.~\ref{eq:gradcam_channel}, we can produce channel-wise class-discriminative activation maps $\{\bm{S}_k\}_{k=1}^K$  for a given class $c$. However, it is challenging to discover meaningful visual concepts from an individual channel, and the channel not that important to the target class might contain noise.
Therefore, we propose to remove the unimportant channels and then use SVD to the rest of the channels to derive common patterns. \YYG{Specifically, from all $K$ class-discriminative activation maps $\{\bm{S}_k\}_{k=1}^K$ of Eq.~\ref{eq:gradcam_channel}, we firstly prioritize and select the top-$P$ maps based on descending gradient values, identifying those most influential for prediction. 
\begin{align}
\label{eq:gradient-selected}
&\{\bm{S}^g_p\}_{p=1}^P = \text{Sort top-}P~\{\bm{S}_k\}_{k=1}^K~\text{by}~\sum_{i=1}^{M} \sum_{j=1}^{N} {(\frac{\partial y^c}{\partial \bm{A}_k})}_{ij} \\
& \hat{\bm{S}}^g=[\hat{\bm{s}}_{1}^T,..,\hat{\bm{s}}_{P}^T]~ \text{where}~\hat{\bm{s}}_{i} = \text{Flatten}[\bm{S}^g_i]
\end{align}
where $\hat{\bm{S}}^g$ denotes the gradient-selected class-discriminative activation map , and $\hat{\bm{s}}_{i}^T$ is the vectorized $\bm{S}_{i}^g$. We should note that  $\hat{\bm{S}}^g$ contains the most significant channels that contribute significantly to the target, and it implies fewer noisy features compared to the original activation map $\bm{A}_k$. Then, to further aggregate the significant factors, {SVD is used to decompose $\hat{\bm{S}}^g$. For vectorized activation maps $\hat{\bm{F}}$, we have :}}

\begin{equation}
\label{eq:decomcam_ossm}
\begin{aligned}
\hat{\bm{F}} = \bm{U}_{[1:Q]}^T \hat{\bm{S}}^g \,\,\, \text{where}\,\,\, \underbrace{\bm{U}\bm{\Sigma} \bm{V}^T = \hat{\bm{S}}^g}_{\text{SVD}},
\end{aligned}
\end{equation}
\noindent where \YYG{$U_{[1:Q]}$ are the left singular vectors of $\emph{top-}Q$ singular values of $\hat{\bm{S}}^g$.} By reconstructing the feature maps with the top $Q$ singular values, we effectively remove noisy features, and also get $Q$ common patterns with ranks. We reshape $\hat{\bm{F}}$ back into ${\bm{F}} \in \mathbb{R}^{Q \times M \times N} $ with original spatial dimensions, and denote $q$-th activation map as $\bm{F}_q \in \mathbb{R}^{M \times N}$. Moreover, we upsample $\bm{F}_q$ into the size of the input and get:

\begin{algorithm}
	\textsl{}\setstretch{1.4}
	\renewcommand{\algorithmicrequire}{\textbf{Input:}}
	\renewcommand{\algorithmicensure}{\textbf{Output:}}
	\caption{DecomCAM}
	\label{alg1}
	\begin{algorithmic}[1]
		\REQUIRE Activation Maps $\{\bm{A}_k\}_{k=1}^K$, Gradient Maps $\{ \frac{\partial y^c}{\partial \bm{A}_k} \}_{k=1}^K$, Number of Channel $K$, Number of Priority Maps $P$, Number of OSSM $Q$, Neural Network $f(\cdot)$, Target Class $c$
       \ENSURE OSSM $\{H_q\}_{q=1}^Q$ and Saliency Map $L_\text{DecomCAM}$
            \\ \textbf{Decomposition Stage}
            \STATE $w_k^g = \frac{1}{{Z}} \sum_{i=1}^{M} \sum_{j=1}^{N} \frac{\partial y^c}{\partial \bm{A}_k} $
            \COMMENT{$\frac{\partial y^c}{\partial \bm{A}_k} \in \mathbb{R}^{M\times N}$}
            \STATE $\bm{S}_k = w_k^g \odot \bm{A}_k$ 
            \COMMENT{$\bm{A}_k \in \mathbb{R}^{M \times N}, \bm{S}_k\in \mathbb{R}^{M \times N}$ }
            \STATE Sort $\bm{S_k}$ by $w_k^g$ in descending order: $\left\{ \bm{S}_1^g,\dots, \bm{S}_K^g \right\}$
            \STATE $\hat{\bm{s}}_k \gets \text{Resize}(\bm{S}_k)$ 
            \COMMENT{ $\hat{\bm{s}}_k \in \mathbb{R}^{1 \times (M \cdot N)}, \bm{S}_k \in \mathbb{R}^{M \times N} $}
            \STATE $\hat{\bm{S}}^g \gets \text{Concat}\left[\hat{\bm{s}}_1^T, \dots, \hat{\bm{s}}_P^T \right]$
            \COMMENT{$\hat{\bm{S}}^g \in \mathbb{R}^{P \times (M \cdot N)}$}
            \STATE  $\bm{U} \bm{\Sigma} \bm{V}^T \gets \text{SVD}(\hat{\bm{S}}^g)$ 
            \COMMENT{ $\bm{U} \in \mathbb{R}^{P\times P}, \bm{V} \in \mathbb{R}^{(M\cdot N) \times (M \cdot N)}$}
            \STATE  $\hat{\bm{F}} \gets \bm{U}^T_{[1:Q]} \hat{\bm{S}}^g $
            \COMMENT{$\hat{\bm{F}} \in \mathbb{R}^{Q \times (M \cdot N)}, \bm{U}^T_{[1:Q]} \in \mathbb{R}^{Q \times P}$}
            \STATE $\bm{F} \gets  \text{Reisze}(\hat{\bm{F}})$
            \COMMENT{$\bm{F} \in \mathbb{R}^{Q \times M \times N}$}
            \STATE $[\bm{H}_1, \dots, \bm{H}_Q] \equiv \bm{H} \gets \left[ \text{s} \circ \text{up} \right] (\bm{F})$  
            \COMMENT{$\bm{H} \in \mathbb{R}^{Q \times H \times W}$} \\
            \COMMENT{\text{s}: Min-Max normalization, \text{up}: Upsampling}
            \\ \textbf{Intergration Stage}
            \FOR{$q \gets 1$ to $Q$}
                \STATE $g(\bm{I}) = \text{GaussianBlur}(\bm{I})$ \COMMENT{$g(\bm{I}) \in \mathbb{R}^{3 \times H \times W}$}
                \STATE $\bm{I}_q^{b} \gets \text{max}(\bm{I} \odot \bm{H}_{q}, \ g(\bm{I}) \odot (1-\bm{H}_{q}))$ \COMMENT{$\bm{I}_q^{b} \in \mathbb{R}^{3 \times H \times W}$}
                \STATE $\Delta\y_q \gets f(\bm{I}_q^{b}) - f(g(I))$ 
                \COMMENT{$\Delta\y_q \in \mathbb{R}^{C}$}
            \ENDFOR
            \STATE $\left[w_1, \dots, w_Q\right] \gets  \text{Softmax}\left(\left[\Delta y_1^c, \dots \Delta y_Q^c\right]\right)$
            \STATE $L_\text{DecomCAM} \gets \sum_{q=1}^Q w_q\bm{H}_{q}$ \COMMENT{$L_\text{DecomCAM} \in \mathbb{R}^{H \times W}$}
		
	\end{algorithmic}  
\end{algorithm}

\begin{equation}
\bm{H}_{q} = [s \circ \text{up}](\bm{F}_{q}),
\end{equation}

\noindent \YYG{where {$\bm{H}_{q} \in \mathbb{R}^{H \times W}$ denotes the sub-saliency maps obtained via the $q$-th singular vectors of the gradient-guided selected class-discriminative activation map} ,  $\text{up}(\cdot)$ denotes the bi-linear interpolation upsampling operation, $s(\cdot)$ denotes a Min--Max Normalization function
that maps each element of input matrix into $\begin{bmatrix}0, 1\end{bmatrix}$.} 
Especially, based on the properties of SVD~\citep{strang2022introduction}, different column vectors in $\hat{\bm{F}}$ are orthogonal. In this case we highlight $\{\bm{H}_{q}\}_{q=1}^Q$ as orthogonal sub-saliency maps (OSSMs). Based on our gradient-guided channels selection and SVD, we produce $Q$ OSSMs for a given class $c$, where $Q \ll K$. 
\vspace{0.5cm}

\noindent \textbf{{Integration.}} A straightforward way to generate final saliency maps is a simple summation of OSSMs like Eq.~\ref{eq:gradcam_fusion}. However, this will lead to false confidence~\citep{bib29}, \ie, activation maps with higher weights might show lower contribution to the network's output. To get rid of false confidence, we propose to use OSSM's impact of the forward passing score on target class, as their weights. The pipeline of integration is shown in Fig.~\ref{fig:overview}.

Given $q$-th OSSM $H_q$ for class $c$, we introduce to create corresponding blurred image $\bm{I}^b_q$ that is sharp in the region activated by $H_q$ and blurred in the remaining region, as shown in  Fig.~\ref{fig:overview} ``Gaussian-blurred image of OSSMs''. We achieve this based on with the following formula:
\begin{equation}
\bm{I}_q^{b} = \text{max}(\bm{I} \odot \bm{H}_{q}, g(\bm{I}) \odot (1-\bm{H}_{q})),
\end{equation}
\noindent \YYG{where $\bm{I}_q^{b}$ denotes the $q$-th blurred image, the Gaussian blur operation $g(\cdot)$ and the element-wise maximum operation $\text{max}(\cdot)$.} Subsequently, we construct a total blurred image $g(I)$ as a reference image and calculating attributing value of the $q$-th OSSM $\bm{H}_{q}$ via comparing their forward inference score:
\begin{equation}
    \Delta\y_q = f(\bm{I}_q^{b}) - f(g(I)),
\end{equation}
\YYG{where $\Delta\y_q \in \mathbb{R}$ denotes the score difference of the blurred image of the $q$-th OSSM $\bm{H}_{q}$ and $f(\bm{I}_q^{b}) \in \mathbb{R}$, $ f(g(I)) \in \mathbb{R}$ denotes the forward inference score of the blurred image of $q$-th OSSM $\bm{I}_q^{b}$ and reference image $g(I)$ , separately.} The final weights $\{w_q\}_{q=1}^Q$ are the softmax normalization over the score difference $\{\Delta y_q^c\}_{q=1}^Q$. Finally, we get saliency maps $L_\text{DecomCAM}$ by a linear combination of weights and OSSMs,


\begin{equation}
\label{eq:decomcam_fusion}
\begin{aligned}
L_\text{DecomCAM} = \sum_{q=1}^Q w_q\bm{H}_{q}.
\end{aligned}
\end{equation}

\subsection{Pseudocode and Application}~\label{subsec:app}
\YYG{Here, we present a pseudocode for our proposed \ours~algorithm in Algorithm \ref{alg1}. For VLM models, the input item $y^c$ can be substituted with the similarity value between the image embedding and the text embedding of the target concept.} Then, to highlight the interpretation for recent VLM-based methods, we conduct all interpretability experiments on CLIP model. CLIP ~\citep{radford2021learning} is distinct in its architecture, featuring dual encoders processing the image and text, respectively. With large-scale pretraining on an extensive dataset of 400 million image-text pairs, CLIP has learned a broad range of visual concepts. Unlike traditional classifiers, CLIP does not directly yield a confidence score for the specific class. Instead, it predicts by evaluating the cosine similarity score between textual prompts and image representations, which is also served as the target vector for interpretability methods \citep{li2022exploring, chefer2021generic}.

\section{Experiment}
\textbf{}
To verify the effectiveness of our DecomCAM, we conduct zero-shot locating interpretation in Sec.~\ref{sec:loc_interp} and causal interpretation in Sec.~\ref{sec:causal_interp}. Moreover, Sec.~\ref{sec:attr_analysis} reveals that our intermediate OSSMs are linked to attributes of objects and can provide a category-level interpretation. The datasets, metrics and implementation details can be found in Sec.~\ref{sec:experimental_setting} and Sec.~\ref{subsec:app}, respectively.

\YYG{
\subsection{Experimental Setup}
In our experiments, we set the default hyperparameters of our proposed method, \ours, as $P=100$ and $Q=10$. Following the methodology akin to the CAM-based approach, we choose the closest valid channel to the CLIP prediction layer, specifically, the final convolutional layer of CLIP's stage-4 architecture, to generate interpretability visualizations. The detailed introduction towards CLIP is included in Supplement \ref{app:CLIP}. To employ CAMs on CLIP, we follow previous work~\citep{chen2022gscorecam} to construct text prompts. Specifically, for the PASCAL-Part dataset, we utilize ``\{class name\} \{part name\}'' (without quotation marks) as the prompt, while for other datasets, we adopt ``\{class name\}'' as the prompt.
}

\subsection{Datasets and Evaluation Metrics}
\label{sec:experimental_setting}
\textbf{Datasets.} 
We conduct experiments on ImageNet-v2~\citep{choe2020evaluating}, Pascal VOC 2012~\citep{bib41}, MSCOCO 2017~\citep{lin2014microsoft}, PartImageNet~\citep{he2022partimagenet} for locating interpretation, Performance-Stratified ImageNet Subsets (PS-ImageNet) for causal interpretation, and PASCAL-Part~\citep{chen2014detect} for providing the attribute-level annotation. A brief introduction of these datasets is as follows:

\textbf{ImageNet-v2}  incorporates additional positional annotations for original 1000-class ImageNet.  We utilize all 10,000 images from its train-fullsup set, also known as val2, for analysis. 
\textbf{MSCOCO 2017} is an 80-class object detection dataset, characterized by images with multiple objects of varying sizes and spatial positions. We employ the entire validation set, encompassing 5,000 images for analysis.
\textbf{PartImageNet} is an extension the ImageNet, providing additional locating annotations for body part of an object for selected images. Our analyses utilize the full suite of 4,598 images from the test set.
\textbf{PS-ImageNet} is a performance-stratified ImageNet subset crafted by ourselves for causal interpretation evaluation. {It re-organizes ImageNet-100~\citep{le2015tiny}} into ten subsets based on 10\% intervals of class-level performance, as gauged by CLIP-ResNet50's classification accuracy. We take 100 images for each class, totaling 10,000 images. The specific stratification result is detailed in Supplement \ref{app:PS-ImageNet}. \textbf{PASCAL VOC 2012} provides 20 visual object classes in realistic scenes for detection. We use the full suite of 5823 images from its validation set.
\textbf{PASCAL-Part} enriches the original PASCAL VOC 2010 dataset with detailed annotations for body parts across various objects, enabling localization tasks at attribute level. We have incorporated all 4,737 training images from this dataset for analysis. 

A more detailed introduction for these datasets is included in Supplementary Material.

\begin{figure*}[tp]
    \centering
     \includegraphics[width=\linewidth]{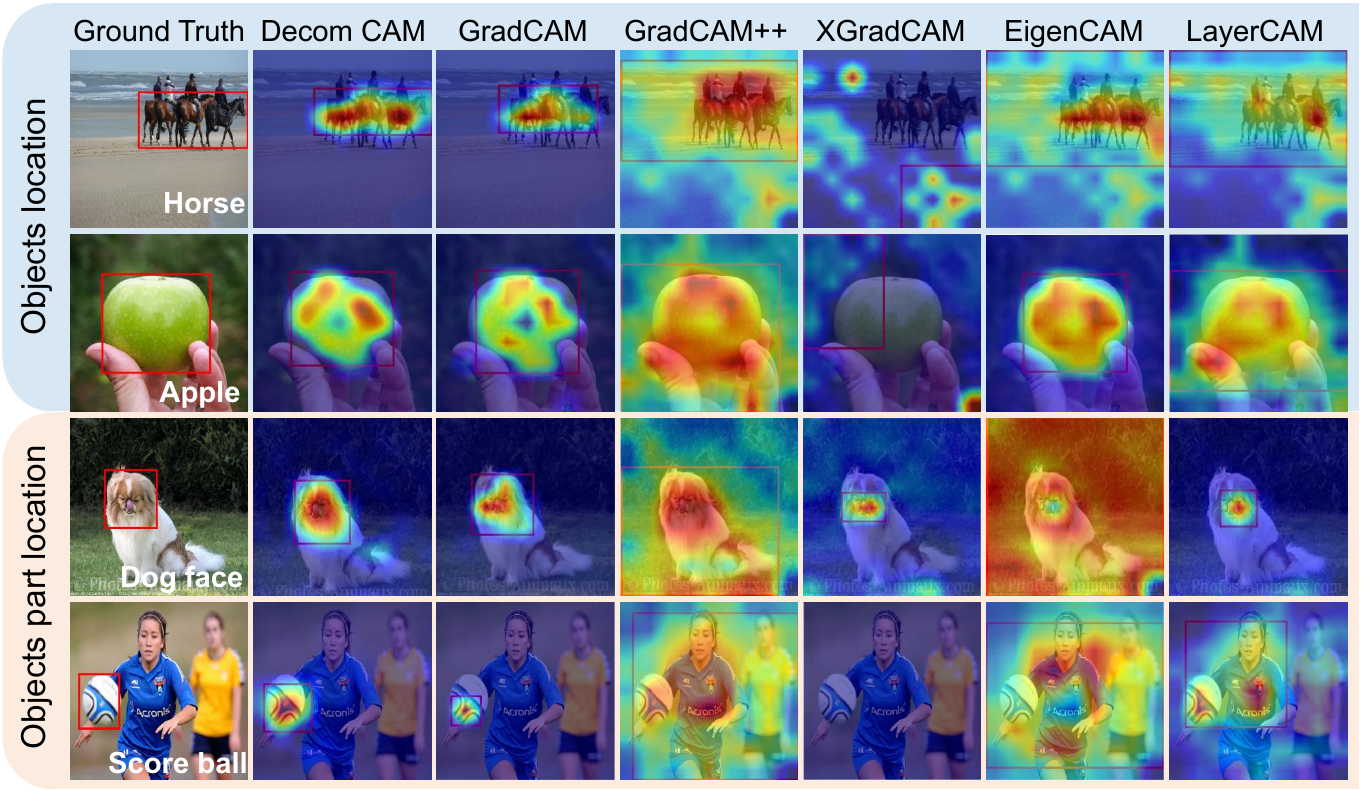}
    \caption{
    Visualization of saliency maps of different gradient-based CAMs. DecomCAM consistently outperforms other methods, producing saliency maps that are more focused on the target concept with less noise. The results are drawn from the CLIP-ResNet50x4 backbone. We generate the predicted bounding boxes following~\cite{chen2022gscorecam}.
    }
    \label{fig:vis_cams}
\end{figure*}

\noindent\textbf{Metrics.}  
For the localization accuracy, we follow~\citep{chen2022gscorecam} and adopt BoxAcc~\citep{gupta2020contrastive} for MSCOCO 2017 and PartImageNet datasets, and MaxBoxAccV2~\citep{choe2020evaluating} for ImageNet-v2. \YYG{We include the calculating process of localization metrics in Supplement~\ref{app:zero-shot detection metrics}}
For the causal interpretation, following~\citep{hesse2021fast, lundberg2020local}, Keep Absolute metric (KAM) and Remove Absolute metric (RAM) are employed to evaluate the faithfulness of CAM outputs.  \YYG{We include the calculating process of causal interpretation metrics in Supplement~\ref{app:causal interpretation metrics}}
For the capability in attribute localization, Pointing Game (PG-ACC)~\citep{selvaraju2017grad} and MaxBoxAccV2 are used.  \YYG{We include the calculating process of attribute localization metrics in Supplement~\ref{app:attribute analysis metric}}.

\subsection{Zero-Shot Locating Interpretation}
\label{sec:loc_interp}

\begin{table*}[t]
\centering
\resizebox{2\columnwidth}{!}{%

\renewcommand{\arraystretch}{1.3}
\begin{tabular}{lcccccc} 
\toprule
 Dataset (Metric)& \multicolumn{2}{c}{ImageNetV2 (MaxBoxAccV2 $\uparrow$)}& \multicolumn{2}{c}{COCO (BoxAcc $\uparrow$)}& \multicolumn{2}{c}{PartImageNet (BoxAcc $\uparrow$)}\\ 

  Methods& RN50$\times$4& RN50$\times$16&RN50$\times$4&RN50$\times$16&RN50$\times$4&RN50$\times$16\\

\midrule

  GradCAM  \cite{selvaraju2017grad} & 32.61& 38.90&9.86&11.59
&5.65&7.72
\\ 
 GradCAM++ \citep{bib24} & 46.23& 44.15
&6.11&5.6
&7.6&6.64
\\ 
 xGradCAM \citep{fu2020axiom} & 18.94& 24.24& 10.68&9.68&2.34&1.77\\
 \midrule
 
  EigenGradCAM \citep{jacobgilpytorchcam} & 2.44& 11.2&0.71&2.24&0.34&3.02\\
 EigenCAM \citep{muhammad2020eigen} & 41.74& 43.61
& 11.48& 11.39
& 11.88&\textbf{11.18}
\\ 
  \ours (ours) & \textbf{57.01}& \textbf{46.49}&\textbf{17.83}&\textbf{13.90}&\textbf{12.24}&{11.17}\\ 
  
  \bottomrule
\end{tabular}
}
\caption{Zero-shot locating experiment results across datasets with varying granularity.  The selected datasets represent distinct challenges: ImageNetV2 for single-object location, COCO for multi-object location, and PartImageNet for object part location.  We benchmark against gradient-based CAMs (GradCAM, GradCAM++, xGradCAM) and decomposition-based CAMs (EigenGradCAM, EigenCAM) with different backbones: CLIP-ResNet50x4 and CLIP-ResNet50x16.}
\label{tab:detection}
\end{table*}

\begin{table}
    \centering
    \small
    \begin{tabular}{lc}
    \toprule
         Method& PG-ACC $\uparrow$\\
    \midrule
         ScoreCAM \citep{bib29} & 67.29 \\
         ScoreCAM++ \citep{bib30}& 66.53 \\
         AbalationCAM \citep{bib36}& 48.31 \\
    \midrule
         GradCAM \citep{selvaraju2017grad}& 82.85 \\
         GradCAM++ \citep{bib24}& 85.75 \\
         WGradCAM \citep{bib35}& 72.99 \\
         \ours (ours) & \textbf{86.86}\\
    \bottomrule
    \end{tabular}
    \caption{Pointing game accuracy comparison on the Pascal-VOC 2012 dataset.  This table presents the PG-ACC of different CAM methods when applied to the CLIP-ResNet50 model. }
    \label{tab:PointingGame}
\end{table}

The zero-shot locating (ZSL) interpretation experiment aims to measure the precision of saliency maps.
Specifically, we locate a single object on ImageNet-v2, multiple objects on MSCOCO 2017, and parts of a single object on PartImageNet. The experimental results are tabulated in Tab.~\ref{tab:detection}.

\textbf{\ours~excels in complex scenario localization with MaxBoxAccV2 and BoxAcc.}
ImageNetV2 contains 1000 object classes, \YYG{which poses challenges due to its diverse range of objects and backgrounds. This means that the interpretability method should perform stably on diverse object classes. We interpret the results of CLIP-ResNet50$\times$4 and CLIP-ResNet50$\times$16 on ImageNetV2. }

For CLIP-ResNet50$\times$4, \ours~achieves an impressive MaxBoxAccV2 score of 57.01\% and outperforms the second-best method GradCAM++ in a large margin (GradCAM++'s 46.23\% vs. Our 57.01\%) . Such notable improvements highlight the effectiveness of \ours ~accurately locating objects from a diverse set of classes.  
\YYG{Despite the class-discriminative interpretability capabilities demonstrated by gradient-based methods like GradCAM++, \ours' decomposition ability extracts key patterns from activation maps to generate saliency maps, which effectively reduces the noises or channels contributes minimally to the target from activation map, resulting in a more stable and outstanding performance across a diverse set of classes.}

To thoroughly evaluate \ours~performance, we further conduct ZSL experiments on MSCOCO 2017, which presents a wide variability of images with diverse target sizes, shapes, and locations.
In this challenging scenario, for CLIP-ResNet50$\times$4, \ours~achieves a remarkable 6.35\% BoxACC improvement  compared to the second-best method (EigenCAM's 11.48\% vs. our 17.83\%) in accurately interpreting model predictions.
\YYG{This improvement can be attributed to \ours's ability to decompose activation maps into multiple OSSMs, allowing it to extract different target positions from feature maps, enhancing its ability to capture multiple objects. Additionally, \ours's denoising ability enables it to generate precise and clear saliency maps for the target concept even in visually cluttered or small target scenes.}

\textbf{\ours~is more reliable in locating fine-grained and important object components.} {As previous works~\citep{zeiler2014visualizing, yu2016visualizing} suggest}, the activation maps of CNNs can locate fine-grained and important object components \wrt objects, such as the dog head \wrt a dog, wheels \wrt a car.

Motivated by this insight, we conducted an object components locating experiment on the PartImageNet dataset \YYG{to verify whether the saliency map can effectively capture these fine-grained and important components.}
Surprisingly, as demonstrated in Tab.~\ref{tab:detection} PartImageNet, existing saliency maps performed poorly except decomposition-based methods (\ie, EigenCAM, and \ours).  For CLIP-ResNet50$\times$4, EigenCAM and \ours~achieve 11.88\% and 12.24\% BoxAcc, respectively. In contrast, other methods are lower than 5.65\% BoxAcc. This observation shows that most existing CAM-based methods tend to generate coarse-grained saliency maps, where the important factors learned by the model might become amalgamated within the saliency map. We also show the visualization of object part localization in objects part location of Fig.~\ref{fig:vis_cams}. 
\YYG{This phenomenon may arise from the presence of redundant features in the numerous channels of the VLM, particularly when interpreting local object features. These channels may contain many redundant features (i.e., those that contribute minimally to the target concept), leading to significant noise in the saliency map generated by existing CAM techniques. However, \ours~overcomes this challenge by explicitly decomposing the activation map, effectively filtering out most of the noise and maintaining excellent performance when explaining local features.

For instance, in the third or fourth row of Fig.~\ref{fig:vis_cams}, existing CAMs exhibit two extremes. In the first scenario, redundant features dominate a significant positive gradient value, leading to the activation of most areas in the image (as observed with GradCAM++ and EigenCAM). Conversely, in the second scenario, redundant features occupy numerous negative gradient values, resulting in insufficient coverage of the ``dog face'' or ``score ball'' area after aggregation (as seen with GradCAM and xGradCAM). In contrast, \ours~effectively extracts factors relevant to the target directly from the activation map for analysis, thereby avoiding interference from redundant features and achieving optimal performance. As shown in the second column, \ours~generates saliency maps precisely located on the local part. Additionally, it's important to note that even though the ``score ball'' is not a component of an object, it occupies only a small fraction of the image. Therefore, it shares the same characteristics as the ``dog face'' cases.}
For CLIP-ResNet50$\times$16, our BoxAcc is slightly lower than EigenCAM's. However, this does not affect our conclusion.

\textbf{\ours~excels in localization with pointing-game test.}
CAMs are commonly used as priors for weakly-supervised localization without requiring a tightly fitting bounding box. \YYG{In contrast to the MaxBoxAccV2 and BoxAcc metrics, the pointing-game test offers a more tolerant evaluation of a saliency map's ability to locate objects. This tolerance makes it especially suitable for assessing the quality of CAM's prior locating performance, where the exact boundary details are less critical.}  Thus, we also conduct the pointing-game test for CLIP-ResNet50 on Pascal VOC 2012 to further assess \ours's capabilities of localization. The PG-ACC comparison is shown in Tab.~\ref{tab:PointingGame}. We can see that \ours~achieves an impressive 86.86\% PG-ACC, highlighting its potential in providing accurate locating priors. 

\begin{figure*}[tp]
    \centering
     \includegraphics[width=\linewidth]{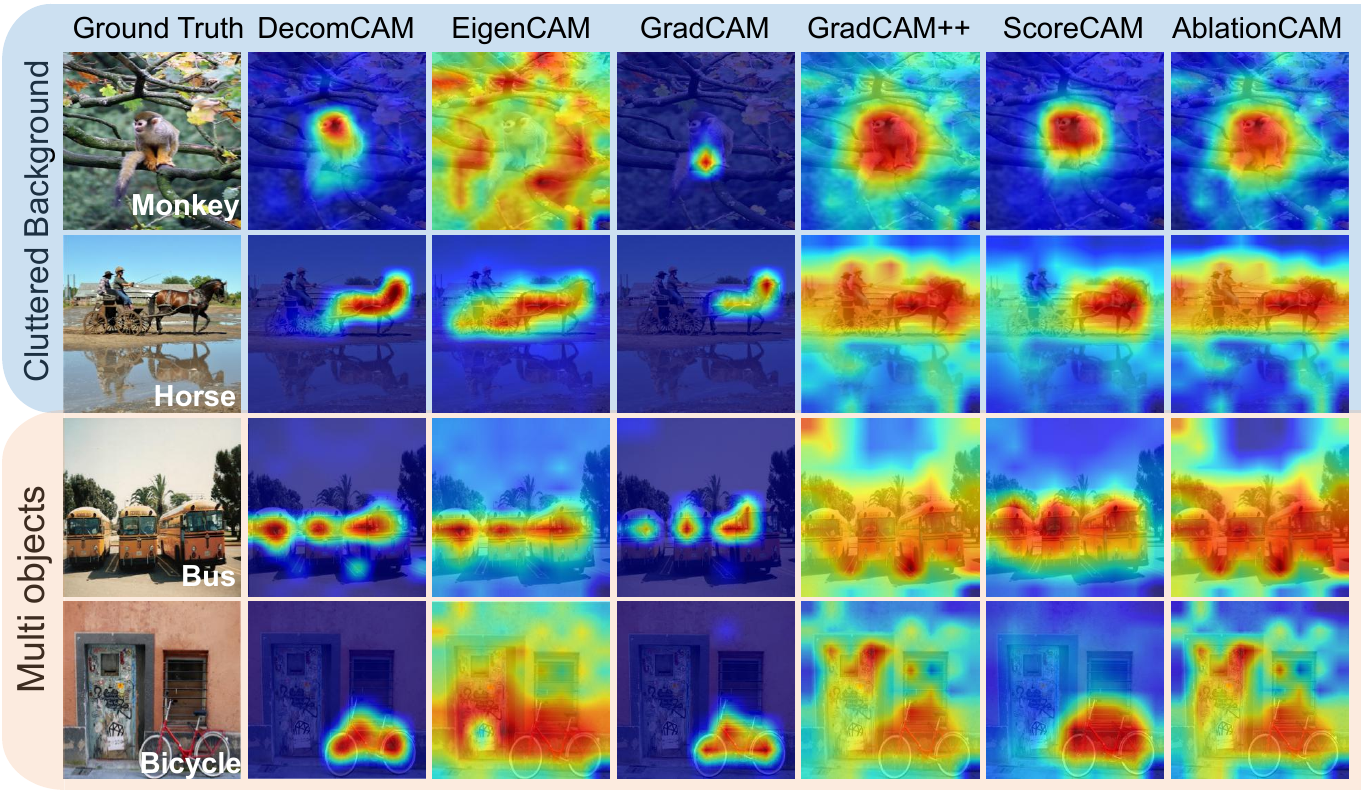}
    \caption{{
    Visualization of saliency maps of different CAM methods. The ResNet50-CLIP is used. \ours~consistently outperforms other methods, showing a more focused and clear saliency map.}}
    \label{fig:vis_cams2}
\end{figure*}

\begin{table*}[t]
\centering
\resizebox{1.7\columnwidth}{!}{%

\renewcommand{\arraystretch}{1.3}
\begin{tabular}{lccccc} 
\toprule
 Method & KAM $\uparrow$ & RAM $\downarrow$ & Overall $\uparrow$ & Runtime $\downarrow$ & Productivity $\uparrow$ \\ \hline

  ScoreCAM \citep{bib29}& 44.54& 15.13&29.41 &45.261 &<1\\ 
  AbalationCAM \citep{bib36}& 37.78& 17.03&20.75 &45.512 &<1\\ 
 ScoreCAM++ \citep{bib30}& 45.19& 14.51&30.68 &17.706 &<1\\ 

\midrule
 GradCAM \citep{selvaraju2017grad}& 47.56& 12.80& 34.76 &\textbf{0.148}&\textbf{234.86}\\ 
  GradCAM++ \citep{bib24}& 48.18& \textbf{11.71}&36.47 &0.179 &203.74\\
  WGradCAM \citep{bib35}& 48.22& 14.83&33.39 &0.150 &222.60\\
   
  \ours~(ours)& \textbf{52.67}& {12.87}&\textbf{39.80} &{0.182}&{218.68}\\ 
\bottomrule

\end{tabular}

}
\caption{Comparative results of the causal interpretability experiment on PS-ImageNet. This table delineates the performance metrics of various interpretability methods tested on the CLIP-ResNet50 model. \ours, with parameters set at $P=100$ and $Q=10$, achieves the optimal balance between interpretability performance and computational speed. 
}
\label{tab:Insertion&Deletion}
\end{table*}

\subsection{Causal Interpretation}
\label{sec:causal_interp}
\begin{figure*}[tp]
    \centering
    \includegraphics[width=\linewidth]{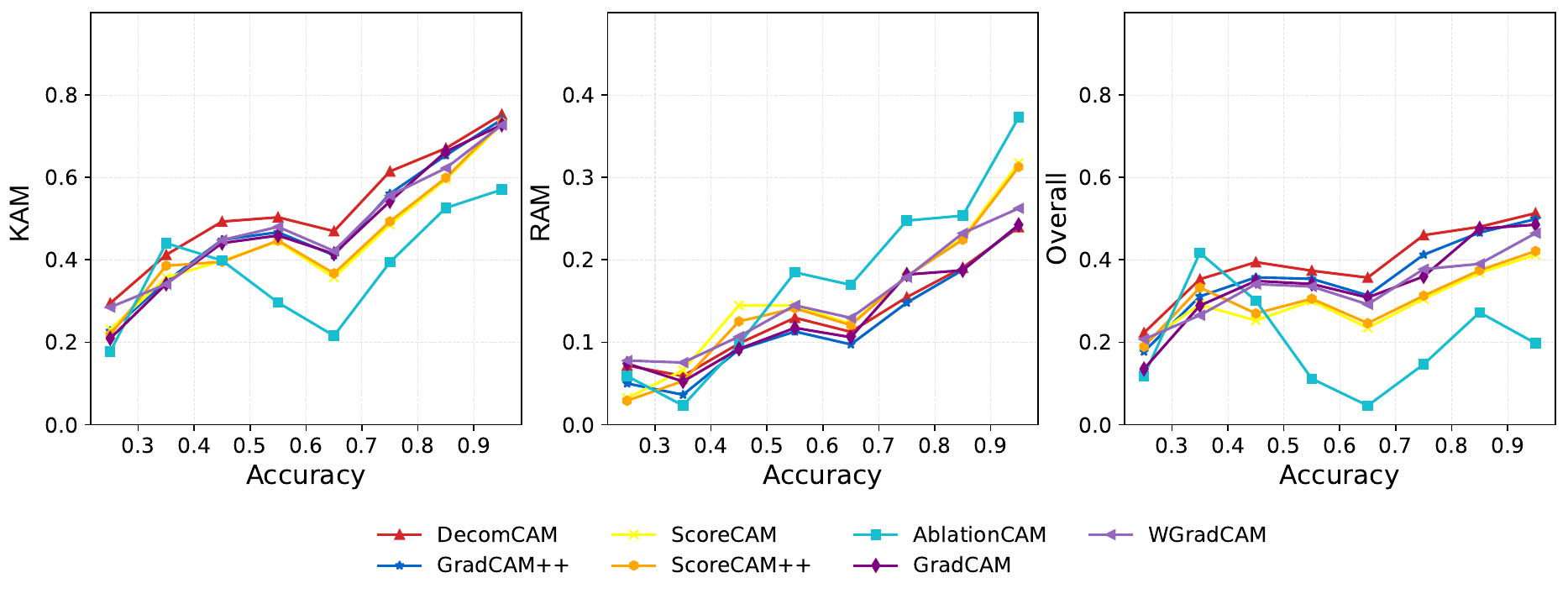}
\caption{
    Causal interpretation performance  on PS-ImageNet. The graph illustrates how each method's interpretability metrics fluctuated with classification performance. Evaluation for various CAMs is performed using CLIP-ResNet50 backbone.}
    \label{fig:interpret}
\end{figure*}


The causal interpretation capacity of CAM refers to its degree of precision in identifying the specific features within an input image that neural networks depend on when predicting the target concept.  

\noindent\textbf{\ours~consistently demonstrates superior performance at all levels of classification accuracy. } \YYG{In this experiment, we assess the interpretability method for the RN50 version of CLIP on PS-ImageNet, where CLIP demonstrates strong discriminative ability across various categories. The challenge lies in determining the efficacy of interpretability methods in elucidating both the categories where CLIP excels and those where it struggles. This requires the precise localization of critical decision regions within the model, which offers insights into both successful and challenging prediction scenarios.}
Fig.~\ref{fig:interpret} shows the performance-interpretability curves derived from PS-ImageNet dataset for CLIP-ResNet50. We can see that most methods, except Ablation-CAM, exhibit similar trends across different levels of classification accuracy. Our approach, \ours, demonstrates superior performance, surpassing other methods across all accuracy strata.  {The outlier performance of Ablation-CAM can be attributed to its methodology of zeroing out activation maps to compute the contribution scores, a technique which may not be ideally suited for large-scale pretrained models like CLIP.} 

Tab.~\ref{tab:Insertion&Deletion} further provides a quantitative analysis. ``Overall'' is defined as the discrepancy between KAM and RAM, \ie, KAM minus RAM. \ours~excels with a notable improvement of $4.49\%$ in KAM and $3.33\%$ in Overall compared to the second-best method. \YYG{We attribute this improvement to the inherent characteristics of \ours, wherein the OSSM can effectively extract the essential factors from the activation map while filtering out the influence of redundant features that contribute minimally to the target concept. As a result, \ours~consistently produces precise saliency maps, and brings a significant improvement.}
Note that Grad-CAM++ achieves lower RAM than ours. We attribute this to the saliency maps of Grad-CAM++ tend to cover larger regions, as show  in  Fig.~\ref{fig:vis_cams}. This inevitably introduces an obvious drop in KAM. 
From a more comprehensive aspect combing KAM and RAM, \ie, Overall score,  \ours~still significantly outperforms Grad-CAM++.

\noindent \textbf{\ours~strikes an optimal balance between interpretability performance and computational speed.} Computational efficiency is crucial for applications. We have calculated the ``Runtime'' of different methods for CLIP-ResNet50 based on the hardware of one A5000 GPU and Intel(R) Core(TM) i9-10900X @ 3.70GHZ CPU.  We further define ``Productivity'' as the ratio of the Overall to Runtime, \ie, Overall divided by Runtime. Tab.~\ref{tab:Insertion&Deletion} indicates that \ours~not only advances the state-of-the-art in causal interpretability but also excels in computational efficiency, achieving the highest productivity index of $218.68$. This demonstrates \ours's superior performance-efficiency trade-off, and potential for real-time applications.

\subsection{Attribute Analysis}
\label{sec:attr_analysis}


\begin{figure}[tp]

    \centering
    \includegraphics[width=\linewidth]{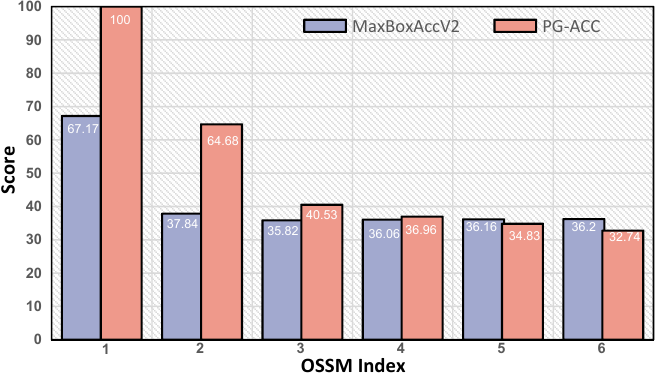}
    \caption{Recall rates of OSSM on PASCAL-Part dataset. This bar chart displays the recall rates of OSSMs \wrt the top-$i$ singular values using CLIP-ResNet50x4 for evaluation. The x-axis represents the index $i$ of the singular values, indicating the rank of importance assigned to each OSSM.}
    \label{fig:attr-ossm}
\end{figure}
\begin{figure}[tp]
    \centering
    \includegraphics[width=1\linewidth]{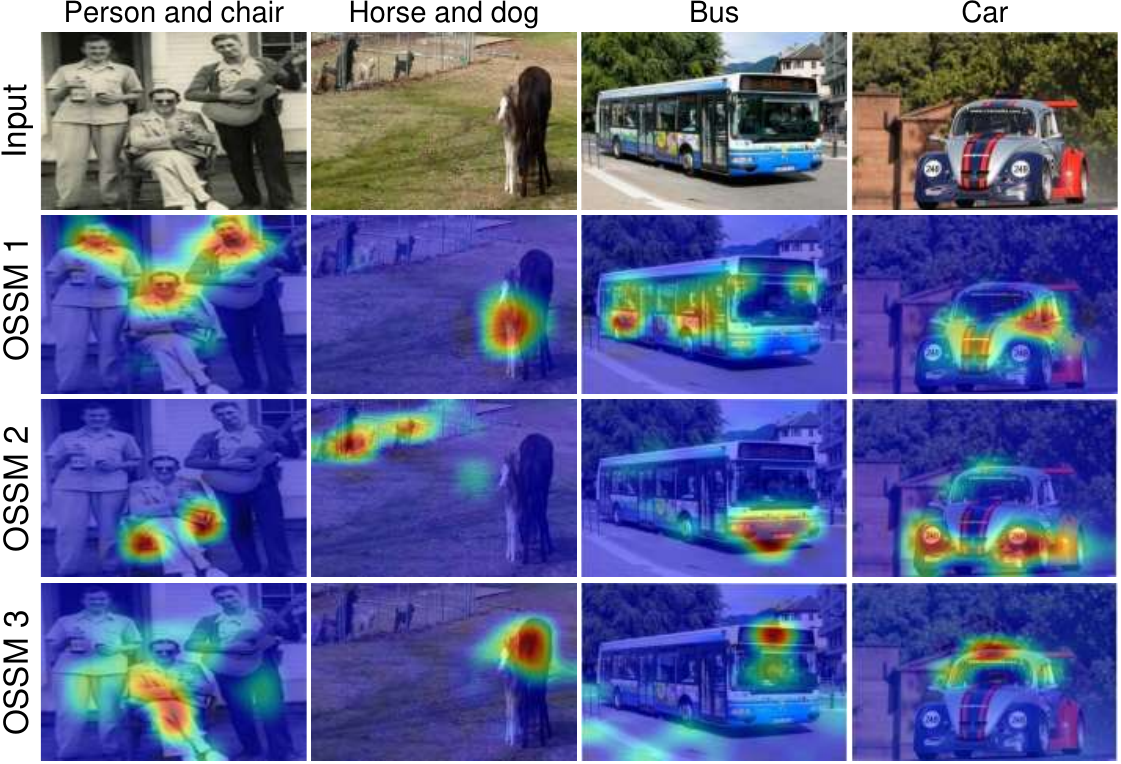}
    \caption{Visual concept localization using OSSMs. This illustration showcases the efficacy of \ours~in isolating and identifying critical visual concepts. For composite concepts ``Person and chair'' and ``Horse and dog'', \ours~effectively decomposes the scene into distinct elements, akin to multi-object localization. In contrast, for singular object concepts  ``Bus'' and ``Car'', \ours~highlights the attributes and differentiates between the various parts of the vehicles. }
    \label{fig:cls_interp}
\end{figure}
\begin{figure}[tp]
    \centering
    \includegraphics[width=\linewidth]{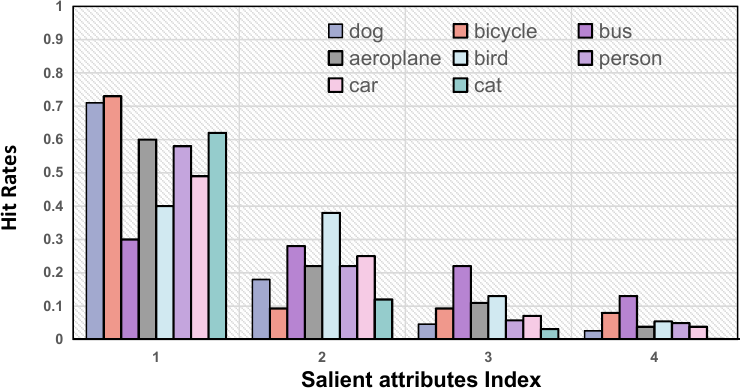}
   \caption{ Hit rates of preset attributes on Pascal-Part dataset. This bar graph depicts the hit rates of the most salient attributes for the classes. The x-axis represents the ranking of attributes based on their hit frequency (index 1 \wrt the top 1 attributes most frequently identified). It is observed that for most classes, the first three attributes are more frequently highlighted by the OSSMs, indicating a concentration of the model's attention on these predominant features. These results are drawn based on CLIP-ResNet50x4.
   }
    \label{fig:attr-pg}
\end{figure}
\begin{figure}[tp]
    \centering
    \includegraphics[width=\linewidth]{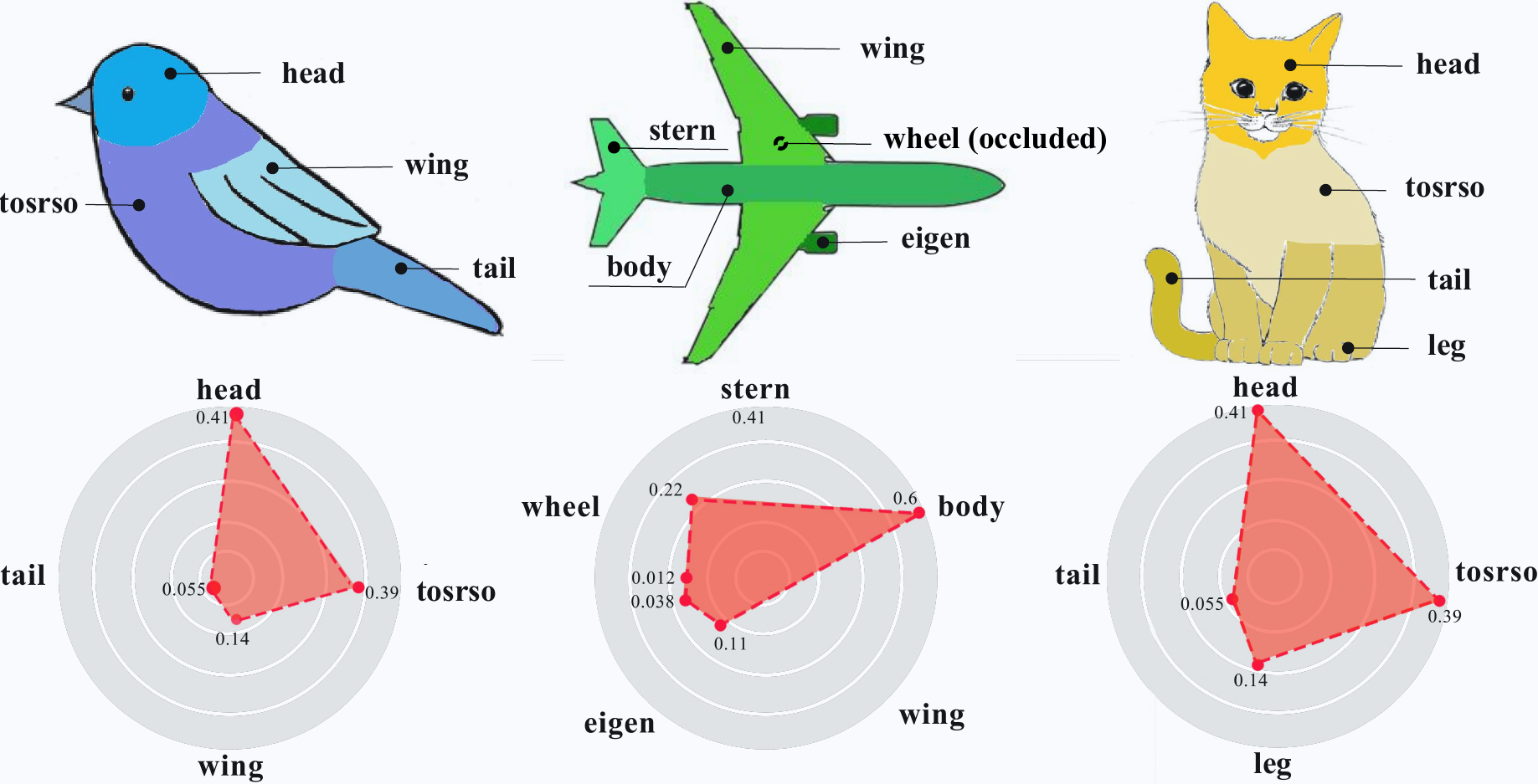}
    \caption{Illustration of the category-level interpretation on Pascal-Part dataset. Here we  showcase the category-level interpretation of ``bird'',  ``plane'', and ``cat''. The experiments are executed on CLIP-ResNet50x4. Each radar chart displays the proportion of attributes, recognized by the model when predicting each category.}
    \label{fig:cls-level-interp}
\end{figure}

\YYG{As revealed by the preceding experiments, \ours~benefits from the orthogonal sub-significance maps (OSSM) generated during the decomposition process, which can extract important factors from the activation map and produce precise saliency maps. An intuitive question arises: Can these factors directly correspond to components of the target concept, such as the wheels, engine, or wings of an airplane?} To reveal the connection between interpretable components of objects, \ie, attributes, and our OSSMs, we further explore attribute-level analysis for our method. We leverage the PASCAL-Part dataset, with its extensive body part annotations, to evaluate the precision of the OSSMs yielded in the decomposition stage.

\noindent \textbf{Evaluation.} \YYG{For evaluation, we undertake a recall-oriented experiment to measure the $\text{HitRate}$ of the OSSM produced by \ours~that can locate any attribute within a sample set.  The accuracy of this localization is determined using MaxBoxAccV2, which evaluates whether OSSMs can pinpoint the specific attribute. The concrete calculating process of $\text{HitRate}$ is included in Supplement \ref{app:attribute analysis metric}. }
The detailed results of this analysis are depicted in Fig.~\ref{fig:attr-ossm}, which illustrates the proportion of attributes each OSSM can reliably identify.

\noindent\textbf{OSSMs can precisely locate the attribute of an object.}  
OSSMs demonstrate remarkable precision in singling out attributes.
Based on Fig.~\ref{fig:attr-ossm}, we can see the OSSM corresponding to the highest singular value achieving a score of 67.17\% for a strict localization match (\ie, MaxBoxAccV2) and 100\% for a relatively tolerant one (\ie, PG-ACC).
This shows OSSMs' success in localizing at least one attribute per image, and indicates that OSSMs associated with larger singular values are more effectively highlighting object attributes.
The OSSMs linked to the third to sixth largest singular values show a relatively similar score of both MaxBoxAccV2 and PG-ACC, which indicates that as singular values decrease, their corresponding OSSMs gradually lose the efficacy of attribute localization.
{As shown in Fig.~\ref{fig:cls_interp},  the OSSM 1 and OSSM 2 show a strong correlation with the object's attributes. However, starting from OSSM 3, we begin to observe an increase in noise, indicating a dilution in the preciseness of attribute localization.  This is aligned with human intuition that attributes contributing more significantly to object recognition are usually expected to be more salient with larger singular values. }

\noindent\textbf{OSSMs facilitate category-level interpretation.} 
\YYG{As demonstrated by the previous experimental results, the top three OSSMs can effectively pinpoint specific attribute regions of objects. This facilitates an intriguing application known as category-level interpretation. Category-level interpretation refers to understanding the model's predictions for a particular category by examining the attributes it relies on and their contributions to the prediction. Attributes are essentially human-defined, so category-level interpretation can reflect the similarity between the model's decision-making criteria and human judgment, thereby assessing the model's interpretability.
}

To investigate this, we conducted a frequency analysis of attribute hit scores, where an attribute is considered ``hit'' when it is identified by an OSSM. The experiment utilized the PASCAL-Part dataset, and the results are depicted in Fig.~\ref{fig:attr-pg}.
Our result reveals an interesting trend that most categories predominantly depend on two or three attributes for recognition. Fig.~\ref{fig:cls-level-interp} unveils our category-level interpretations for categories ``cat'', ``plane'', and ``bird'', illustrating the attributes that the model leverages for decision-making.  For instance, the ``aeroplane'' category is most frequently associated with the ``body'' of the plane , and the ``ster'' of the plane. Specifically, ``body'' and ``ster'' offer 0.6 and 0.22 hit rates, respectively, while ``wings'',``engine'' and ``wheel'' offer lower hit rates. 
\YYG{This observation underscores the capability of OSSMs to provide insights into the decision-making process of the model at the category level, by revealing which attributes contribute most significantly to predictions.}


\section{Discussion and Conclusion}
\noindent \textbf{Enhancing interpretability through decomposition.}  Existing CAM techniques rely on a simple and direct aggregation approach by summing activation maps to highlight channels contributing most significantly to the target, which may result in saliency maps with considerable noise. Differently, \ours~can analyze the correlation between each channel and extract dominant principal components using Singular Value Decomposition (SVD). Compared with the aggregation scheme in CAM techniques, SVD can effectively decompose the intrinsic elements from activation maps while reduce the noise effect. 

The superiority of \ours's precise saliency map lies in the following two reasons:
1) During our decomposition process, channels with larger singular values contributing dominantly to the target interpretation are aggregated into components., By picking up these components, our method can reduce the noise effect from the channels of smaller singular values. Consequently, by selecting the decomposed components with larger singular values, DecomCAM can effectively filter out channels of noisy information and capture the significant factors contributing most towards the target interpretation. 
2) Moreover, the orthogonality of the singular vectors further aids in disentangling the significant factors from the class-discriminative activation maps. Such property allows for a clearer separation of meaningful information from noise, facilitating a more detailed and fine-grained interpretation into the model's decision-making process.

Furthermore, the capability of DecomCAM to extract representative channel-wise patterns may be the key to facilitating fine-grained interpretation. As revealed by Chen and Zhong \citep{bib30}, only a few channels are responsible for interpreting the target concept, and the remaining channels often contain redundant features or noise irrelevant to the target concept. Our zero-shot detection experiment on Pascal-Part also supports this observation. Except for decomposed-based methods, most CAM methods generate activation maps with significant noise when interpreting the local components of objects.

\noindent \textbf{Category-level interpretation.} The application of interpretability methods is an intriguing topic. During our research, one of the most interesting findings was that OSSMs generated by DecomCAM not only theoretically contain the most information from the activation map but also practically reveal the local components of the target concept. This characteristic facilitates a novel application for CAMs, \ie, category-level interpretation. Category-level interpretation aims to attribute the deep learning model's recognition of a category to several attributes, and quantify the contributions of each attribute. As shown in Fig. \ref{fig:cls-level-interp}, We demonstrated how the deep learning model relies on the attributes of each category to make predictions. Category-level interpretation is meaningful because humans can evaluate the model's interpretability and generalization ability by assessing its recognition criteria (i.e., which attributes it mainly relies on for recognition).

\noindent \textbf{Ethical considerations and model transparency.} Moreover, category-level interpretation has broader implications beyond model evaluation. It can also enhance the model transparency and assist developers in identifying whether a deep learning model relies on potentially unethical or biased features for recognition tasks. For instance, it can help detect if a model excessively depends on gender-discriminative features for occupation recognition, thereby promoting fairness and ethical considerations in AI systems.

\noindent \textbf{Limitation and future work.} Although DecomCAM demonstrates promising results across six benchmarks, there are still some limitations can be further explored. Firstly, the decomposition process aims to extract channel-wise common patterns, but it could be enhanced to distill pixel-wise common patterns. This refinement could effectively reduce noise in non-dominant activation regions within channels, leading to clearer and more accurate saliency maps. Secondly, the OSSMs extracted by DecomCAM currently reflect the components of the target concept only for individual samples. Extending this decomposition approach to all samples within the same category in a dataset would be an intriguing direction for further investigation. In the future, we will continue to investigate the use of decomposition and discover interpretable components learned by deep models.

Overall, DecomCAM exhibits a substantial improvement to the interpretation of deep learning model. By providing insights into model decision-making processes and facilitating category-level interpretation, DecomCAM contributes to the development of more transparent, accountable, and ethically sound AI systems.

\section{Acknowledgment}
The work was supported by the National Key Research and Development Program of China (Grant No. 2023YFC3300029), National Natural Science Foundation of China (No. 72001213, No. 12201024). This research was also supported by Zhejiang Provincial Natural Science Foundation of China (Grant No. LD24F020007), Beijing Natural Science Foundation L223024, "One Thousand Plan" projects in Jiangxi Province (Jxsg2023102268), Beijing Municipal Science and Technology Commission, Administrative Commission of Zhongguancun Science Park Grant (No.Z231100005923035), Taiyuan City "Double hundred Research action" (2024TYJB0127).

\printcredits

\bibliographystyle{cas-model2-names}

\bibliography{mybibfile}

\clearpage

\appendix

\YYG{
\section{CLIP Networks}
\label{app:CLIP}

\subsection{Models}
We conduct all our experiments based on the CLIP~\citep{radford2021learning} model. 
Specifically, we employ RN50x16 and RN50x4 for zero-shot detection experiments as they provide the best performance among the convolutional-based variants. For causal interpretation experiments, we utilize RN50, whose recognition performance demonstrates good discriminative ability for both simple and challenging samples, thus enabling a comprehensive evaluation of interpretability methods. In the attribute analysis experiment, we opt for RN50x4 to analyze a model with relatively balanced performance across different attributes, to ensure the applicability of our findings.

\subsection{Target Layers for Visualization}
Similar to the approach employed in the CAM-based family, for interpretability purposes, we select the valid channel closest to the CLIP prediction layer, specifically, the last bottleneck of stage 4 during inference. This corresponds to the final convolutional layer of the image encoder in CLIP. RN50 includes 2048 channels with a spatial dimension of $7 \times 7$, RN50x4 comprises 2560 channels with a spatial dimension of $9 \times 9$, and RN50x16 consists of 3072 channels with a spatial dimension of $12 \times 12$.
}

\YYG{
\section{Zero-shot Detection Experiment}
}

\YYG{
\subsection{Experimental setup}
The Zero-shot Detection experiment involves three datasets with significant differences and distinct challenges, necessitating adjustments to the hyperparameters of DecomCAM for model interpretation. These primary hyperparameters involves: the number of priority maps, denoted as $P$, and the number of orthogonal sub-saliency maps (OSSM), denoted as $Q$. For the ImageNetV2 dataset, characterized by single-object images where the target occupies a significant portion, we set $P=500$ and $Q=1$. The COCO dataset comprises complex scenes with multiple objects, potentially leading to a higher number of object components within activation maps. Therefore, we set $P=500$ and $Q=20$. In contrast, the Part-ImageNet dataset requires fine-grained interpretability to locate components accurately. Hence, we set $P=1000$ to ensure comprehensive coverage of all components and $Q=10$.
}

\YYG{
\subsection{Zero-shot Detection Metric}
\label{app:zero-shot detection metrics}
\noindent \textbf{BoxAcc~\citep{gupta2020contrastive}:} Although CAMs produce saliency maps that can function as unsupervised object detectors \citep{bib24, wagner2019interpretable, Bi-CAM}, determining the detection bounding box remains a challenge. The BoxAcc metric offers an intuitive solution by treating the saliency map as a probability map indicating the presence of the target concept at various positions in the image. By applying a threshold to filter out regions with low confidence and deriving detection boxes based on the minimum bounding rectangle of the remaining areas, BoxAcc provides a robust measure of model performance. The mathematical formula of this metric follows:
\begin{equation}
BoxAcc_{\tau, \delta}(h,\mathbf{B}) :=\frac{1}{M}\sum_{m} \mathbf{1}_{IoU(box(h, \tau )_m, B_m)\ge \delta },
\end{equation}
\noindent where $\tau \in [0,1]$ represents the binarization threshold, $h$ denotes the saliency generated by the model, $box(h,\tau)$ signifies the tightest box around the largest-area connected component of the binarized saliency with threshold $\tau$, $B$ corresponds to the ground truth box, $m \in M$ is the box index, and $\delta$ signifies the IoU threshold, with $\delta=0.5$ utilized in our experiments.

\noindent \textbf{MaxBoxAccV2~\citep{choe2020evaluating}}: This metric follows the same motivation as BoxAcc but conducts a more detailed and accurate search of the two hyperparameters $\tau$ and $\sigma$. It is mathematically defined as:
\begin{equation}
MaxBoxAcc_{\delta}{(h,\mathbf{B})}:= \frac{1}{3}\sum _\delta \max _\tau \big (BoxAcc_{\tau, \delta}(h,\mathbf{B}) \big ),
\end{equation}
where $h$ denotes the saliency map, $B$ denotes the ground truth box, $\delta \in \{0.3, 0.5, 0.7\}$, $\tau \in [0:0.05:0.95]$ (Here, $\tau$ represents the binarization threshold for converting saliency maps into binary masks, with an initial value of 0, a step size of 0.05, and an end value of 0.95.). We adopt the default hyper parameters of this from \cite{choe2020evaluating}.
}
\YYG{
\subsection{Zero-shot Detection Benchmark \& Challenge}
\label{app:zero-shot detection benchmark}
\noindent \textbf{ImageNet-v2}~\citep{choe2020evaluating}: ImageNet-v2 is a single-object detection dataset, which poses challenges in object detection due to its diverse range of objects and backgrounds. Unlike the original ImageNet dataset, which focuses on object classification. ImageNet-v2 introduces variations and complexities that closely resemble real-world scenarios. Despite this dataset contains solely single object in the image, it includes variations in object scales, diverse backgrounds and lighting conditions. Additionally, the presence of similar-looking objects and subtle differences between object classes further complicates object detection tasks.

\noindent \textbf{2017 MS COCO}~\citep{lin2014microsoft}: MS COCO is a multi-object detection dataset, which presents significant challenges for object detection tasks due to its complexity and diversity. The dataset contains images with multiple objects of varying sizes, shapes, and poses, often appearing in cluttered scenes with complex backgrounds. Furthermore, the dataset includes a wide range of object categories, including both common and rare classes, requiring models to generalize well across diverse object classes and variations. Overall, the complexity and diversity of objects, along with the presence of occlusions and cluttered scenes, make object detection on the COCO dataset a challenging task for computer vision algorithms.


\noindent \textbf{Part-ImageNet}~\citep{he2022partimagenet}: Part-ImageNet is a variant of ImageNet where selected images are further annotated at the part level, incorporating a hierarchical labeling system. For instance, within the ``bird'' category, birds are subdivided into parts such as ``head'', ``body'', ``wings'', ``legs'', and ``tail''. The dataset comprises 24,000 images spaning 11 super-categories. The object detection task within Part-ImageNet is notably challenging, demanding precise attribute methods to capture the model's attention for each part of the object. Our evaluation is conducted using all 4,598 images in the test set.
}

\YYG{
\section{Causal Interpretation Experiment}
}

\YYG{\subsection{Experimental setup}
For the causal interpretation experiment, we maintain the default hyperparameter settings of \ours~($P=100$, $Q=10$). We utilize the RN50 version of CLIP for these experiments. The RN50 model has demonstrated balanced discriminative ability across both simple and challenging samples~\citep{radford2021learning}, making it suitable for conducting a comprehensive evaluation of interpretability methods. 

\GRT{

\subsection{Causal Interpretation Metrics}
\label{app:causal interpretation metrics}
\noindent \textbf{KAM and RAM Metrics.} The Keep Absolute mask (KAM) metric and Remove Absolute mask (RAM) metric are widely employed \citep{bib30, bib29, petsiuk2018rise, bib35}:

 In RAM, high-value pixels in the activation map are progressively replaced with the mean value of the input image (effectively blurring the input clear image), resulting in a decrease in the model's output class score. The RAM score is obtained by gradually replacing the highest scoring fraction with the mean value and computing the area under the output score curve of the target class, with lower values indicating better interpretability.

 In KAM, the input image starts with all pixels set to the mean value, and the activation map's highest scoring fraction is gradually added (gradually restoring the input image to its original clear state), resulting in an increase in the output score of the target class. The KAM score is determined by the resulting area under the output score curve of the target class, with higher values indicating better interpretability.

 An overall metric, calculated as the difference between KAM and RAM scores, provides a comprehensive evaluation. However, the reliability of these scores may be influenced by the model's prediction accuracy, particularly in cases of erroneous predictions during testing.

}

 \textbf{Performance-stratified Evaluation of CAMs}: Although KAM and RAM metrics indicate the performance of interpretability methods in explaining predictions of deep learning models, they overlook a crucial fact, \ie, the classes where the interpretability model performs poorly may be more meaningful for guiding actual model corrections compared to those where it performs well. Therefore, we stratify the dataset based on the performance of the interpreted model on each subset (e.g., PS-ImageNet as shown in Supplement \ref{app:causal interpretation benchmark}). Subsequently, we calculate KAM and RAM values separately for each subset and compute the AUC of the KAM and RAM values across different subsets to draw a more comprehensive results.

}

\YYG{
\subsection{Causal Interpretation Benchmark \& Challenge}
\label{app:causal interpretation benchmark}

\noindent \textbf{PS-ImageNet}: 
To facilitate a comprehensive evaluation of interpretability across varying levels of model performance, we introduce the PS-ImageNet dataset. This dataset is a compilation of ten subsets, each representing a performance tier stratified by a 10

The challenge lies in determining whether interpretability methods can effectively explain both the categories where CLIP performs well and those where it performs poorly. This entails the precise localization of key decision areas within the model, providing insights into both successful and challenging prediction scenarios.

}

\subsection{The specific performance-stratified results of PS-ImageNet dataset}
\label{app:PS-ImageNet}

\noindent Subset1: gordon setter, rock beauty, ashcan, cocktail shaker, hair slide, ear

\noindent Subset2: tibetan mastiff, file, nematode, african hunting dog, ibizan hound

\noindent Subset3: arctic fox, worm fence, black footed ferret

\noindent Subset4: komondor, fire screen, tobacco shop, poncho

\noindent Subset5: chime, pencil box, solar dish, tile roof, hourglass, carton, combination lock, miniskirt

\noindent Subset6: house finch, harvestman, boxer, aircraft carrier, frying pan, reel, stage, crate, mixing bowl, theater curtain, rhinoceros beetle

\noindent Subset7: green mamba, dugong, saluki, newfoundland, miniature poodle, wok, malamute

\noindent Subset8: french bulldog, barrel, clog, oboe, parallel bars, upright, street sign, vase, coral reef

\noindent Subset9: triceratops, jellyfish, Walker hound, three toed sloth, beer bottle, dishrag, holster, lipstick, organ, photocopier, prayer rug, snorkel, unicycle, hotdog, cliff, bolete, king crab,  golden retriever, dalmatian, ant, bookshop, cuirass, electric guitar, meerkat, cannon, iPod

\noindent Subset10: robin, toucan, ladybug, carousel, dome, slot, spider web, tank, consomme, orange, lion, school bus, scoreboard, trifle, goose, white wolf, catamaran, garbage truck, missile

\section{Attribute Analysis Experiment}

\YYG{
\subsection{Experimental setup}
Here, we maintain the default hyperparameter settings of \ours~($P=100$, $Q=10$). We utilize the RN50x4 version of CLIP for this experiment.
}

\GRT{
\subsection{Attribute Analysis Metrics}
\label{app:attribute analysis metric}
\noindent \textbf{Hit Rate}: This metric essentially measures the frequency of the OSSM produced by \ours~that can locate any attribute within a sample set.
\begin{equation}
HitRate:=\sum_{\mathbf{a} \in \mathbb{A} }MaxBoxAcc_{\sigma}({h,\mathbf{a}}),
\end{equation}
where $\mathbb{A}$ represents the set of ground truth attributes for the samples, $\mathbf{a}$ denotes the set of ground truth boxes for attribute $a$, and $h$ represents the sub-saliency map corresponding to the OSSM.

}
\YYG{
\subsection{Attribute Analysis Benchmark \& Challenge}
\label{app:attribute analysis benchmark}
\noindent \textbf{Pascal-Part}~\citep{chen2014detect}: 
The PASCAL-Part dataset is an additional annotation dataset complementing PASCAL VOC 2010. It provides segmentation masks for various body parts of each object The dataset includes 20 categories, which can be categorized into four super-categories: Person, Animal, Vehicle, and Indoor. The training and validation sets comprise 10,103 images, Although the test set consists of 9,637 images. This dataset can be employed for animal part detection and general part detection tasks. 

Our evaluation is conducted using the training set of PASCAL-Part. The primary objective of this experiment is to verify whether the Orthogonal Sub-Saliency Maps (OSSMs) generated by DecomCAM effectively correspond to the components (attributes) of objects. The key challenge lies in the ability of DecomCAM to uncover semantic components from the intermediate activation map of the deep learning model.


}
\end{document}